\documentclass{article}

\usepackage{microtype}
\usepackage{graphicx}
\usepackage{subcaption}
\usepackage{booktabs} 

\usepackage{hyperref}
\usepackage{float}
\graphicspath{{figures/}} 

\usepackage{arxiv}
\usepackage[numbers]{natbib}

\usepackage{amsmath}
\usepackage{amssymb}
\usepackage{mathtools}
\usepackage{amsthm}

\usepackage[capitalize,noabbrev]{cleveref}

\theoremstyle{plain}
\newtheorem{theorem}{Theorem}[section]
\newtheorem{proposition}[theorem]{Proposition}

\theoremstyle{definition}

\theoremstyle{remark}

\usepackage[disable,textsize=tiny]{todonotes}

\title{Loss-Conditional PINNs for Parametric PDE Families}

\author{%
  Anna Lazareva \\
  Faculty of Computer Science \\
  HSE University, Moscow, Russia \\
  \texttt{annelzrv@gmail.com} \\
  \And
  Alexander Tarakanov \\
  VK and HSE University \\
  Moscow, Russia \\
  \texttt{atarakanov@hse.ru} \\
}

\begin{document}

\maketitle

\begin{abstract}

Physics-informed neural networks (PINNs) approximate solutions of
ODEs and PDEs by minimising a weighted combination of residual,
boundary, initial, and data losses. Their performance, however, is
often dominated by the choice of loss weights: a poor weighting may
lead training to a degenerate solution in which one physical constraint
is satisfied while another is ignored. Existing methods therefore seek
to select or adapt a single good set of weights. We take a different
view: instead of tuning one weight vector, we explore the entire weight
space during training.

We introduce LC-PINN, which adapts the loss-conditional training of
\citet{dosovitskiy2020} to the PDE-residual setting: the conditioning
vector---either the loss weights or a scalar physical coefficient---is
treated as a network input and sampled from a simple prior at every
optimisation step. This turns PINN training into learning a continuous
family of solutions indexed by that vector, while requiring no
solver-generated paired data. In this sense, LC-PINN lies between
classical PINNs and operator-learning methods: it remains fully
physics-informed, but amortises training over a parametric family. Our
contribution is not the loss-conditional construction itself, but its
extension to PINNs, the unification of the loss-weight and
parametric-coefficient regimes under one architecture (concatenation
for loss weights, FiLM for coefficients), and a fixed-quadrature L-BFGS
finishing protocol that makes the parametric-coefficient regime trainable.

We give a $\lambda$-invariance result characterising the conditional
optimum and study LC-PINN empirically on parametric Helmholtz,
Schrödinger, viscous Burgers, and Buckley--Leverett equations. Across
these problems a single LC-PINN matches or improves retrained per-weight
PINN baselines while parameterising the full family in one model, at a
total training cost that amortises favourably against per-instance
retraining.

\end{abstract}

\section{Introduction}
\label{sec:intro}

Physics-Informed Neural Networks \citep[PINNs;][]{raissi2019}
approximate solutions of differential equations by minimizing a
weighted combination of PDE residual, boundary, initial, and
(optional) data losses. Although highly flexible, PINNs are often
extremely sensitive to the choice of loss weights and physical
parameters. Poorly balanced objectives may lead optimization toward
solutions that satisfy one physical constraint while largely ignoring
others. As a result, practical PINN workflows frequently rely on
careful hyperparameter tuning, adaptive weighting schedules, or even
retraining for every parameter configuration of interest.

This paper addresses this limitation through a different perspective.
Instead of searching for a single optimal weighting configuration, we
train a single conditional PINN across an entire family of loss
weights or PDE parameters. The key idea is to treat the conditioning
variable $\lambda$ as part of the network input, transforming PINN
training into learning a continuous family of solutions indexed by
$\lambda$.

The resulting model, which we call LC-PINN, occupies an intermediate
position between classical PINNs and operator-learning methods. Like
neural operators, LC-PINN amortizes computation across a family of PDE
instances. Unlike operator-learning approaches such as FNO
\citep{li2020fno} or DeepONet \citep{lu2021deeponet}, however,
LC-PINN remains fully residual-based and does not require
solver-generated paired datasets.

Conceptually, the proposed framework rethinks loss balancing in
PINNs. Conventional adaptive-weight methods attempt to identify a
single good scalarization of the multi-objective training loss.
LC-PINN instead explores the geometry of the entire loss-weight space
during training. In this sense, loss weighting becomes not merely a
hyperparameter-tuning problem, but a parametric learning problem.

\paragraph{Contributions.}
The loss-conditional construction is due to \citet{dosovitskiy2020};
conditioning a PINN on a physical parameter has also been done before
\citep{arthurs2021}. Our contribution is to bring these ideas together
for PINNs and to study the result carefully:
\begin{itemize}\setlength{\itemsep}{0pt}
   \item We adapt loss-conditional training to the PDE-residual setting
         and unify two regimes---loss-weight conditioning (concatenation)
         and parametric-coefficient conditioning (FiLM)---under one
         architecture, with a fixed-quadrature L-BFGS finishing protocol
         that makes the parametric-coefficient regime trainable.

   \item We give a $\lambda$-invariance result showing that global
         minimizers of the conditional objective recover residual
         minimizers across the parameter family, and reconcile it with
         the $K$-shot averaging used at inference via a per-sample
         variance analysis (\Cref{sec:k-shot}).

   \item We give a local optimization result showing that, in a
         quadratic model, conditioning supplies additional descent
         directions. We are explicit that this is an instantaneous,
         local statement---a consistency check, not an explanation of
         the order-of-magnitude empirical gains, which arise from
         cumulative smooth-in-$\lambda$ parameter sharing.

   \item We evaluate LC-PINN on parametric Helmholtz, Schrödinger,
         Burgers, and Buckley--Leverett equations against strong adaptive
         baselines (SA-PINN, ReLoBRaLo, Causal-PINN, PI-DeepONet),
         reporting failure cases honestly and quantifying the
         amortisation crossover per family.
\end{itemize}

\section{Related work}
\label{sec:related}

\paragraph{Adaptive loss weighting in PINNs.}
The fragility of the composite-loss objective in the original PINN
formulation \citep{raissi2019} has motivated a line of methods that
adapt the residual / boundary / data weights during training.
Self-adaptive PINN (SA-PINN) of \citet{mcclenny2020} attaches a
trainable per-collocation-point weight that is updated by ascent on
the residual, yielding an effective focusing of the network on
hard-to-fit regions; their ``points-with-weights'' formulation is
essentially a soft hard-example-mining for the residual. ReLoBRaLo
\citep{bischof2021} re-balances loss-term weights from running
ratios of the absolute loss values, smoothed by an exponential moving
average. Causal-PINN \citep{wang2024} weights the residual by a
time-causal mask so the network does not collapse onto late-time
dynamics before fitting the initial-time region. These all produce
\emph{one} solution for \emph{one} (adaptively-found) weighting and
would need to be retrained at every new weighting the practitioner
wishes to inspect.

\paragraph{Operator learning.}
A parallel line of work has built neural operators that map a
function-valued input (boundary condition, forcing term, initial
condition) to a function-valued solution. The Fourier Neural Operator
(FNO) \citep{li2020fno} parameterises the operator in spectral space:
linear layers on the lowest Fourier modes, plus a residual
$1\!\times\!1$ convolution. DeepONet \citep{lu2021deeponet} factorises
the operator into a branch network (encoding the input function at
sensor points) and a trunk network (encoding the spatial coordinate),
recombined by an inner product. Both methods amortise across a family
of problems but require a precomputed dataset of (input, solution)
pairs from a classical solver---they are supervised learners in
function space. The PINN residual itself plays no role in their
training. PI-DeepONet \citep{wang2021pi-deeponet} re-introduces the
residual but stays inside the operator-learning framing, treating the
input function as the parameter rather than a scalar coefficient.
Our LC-PINN sits between these two camps: it amortises like
operator learning but uses the residual loss like a PINN.

\paragraph{Loss-conditional networks.}
\citet{dosovitskiy2020} introduced the loss-conditional networks idea
in image generation: a single generator, conditioned on the loss
weights of a multi-objective composite reconstruction loss, produces
images that interpolate the tradeoff surface (sharpness vs.\ perceptual
fidelity vs.\ adversarial realism). The network is trained by sampling
the weight vector from a fixed prior at every step and applying the
freshly-sampled weights to the per-instance reconstruction loss.
LC-PINN takes the same conditional-on-weights idea and applies it to
the PDE-residual setting; the formal $\lambda$-invariance result
(\Cref{sec:theorem}) is, to our knowledge, the first analysis of the
optimum-set structure of this construction in the residual-loss
case.

\paragraph{Hypernetworks and meta-learning for PINNs.}
A separate amortisation strategy uses hypernetworks
\citep{ha2017hypernetworks} or MAML-style meta-learning
\citep{finn2017maml} to produce per-instance PINN weights. PI-MAML
\citep{liu2022pi-maml} demonstrates the meta-learning angle on
parametric PDEs. These methods amortise at the level of network
weights rather than a network input. They typically require a training
distribution of full PDE \emph{tasks} and a per-task inner-loop
adaptation step at test time; LC-PINN trains a single network with no
inner loop and produces predictions for every $\lambda$ in one forward
pass.

\paragraph{Conditioning PINNs on physical parameters.}
Conditioning a PINN directly on a physical parameter is not itself new.
Most directly, \citet{arthurs2021} train a PINN to aggregate and
interpolate parametric solutions of the Navier--Stokes equations,
sampling the physical parameter (domain shape, boundary conditions)
during training via an active-learning loop so that a single network
covers a region of parameter space. Our parametric-coefficient mode is
closest to this line of work; we differ in (i) conditioning on a
low-dimensional scalar coefficient through FiLM hidden-layer modulation
rather than the network input, (ii) the fixed-quadrature L-BFGS
finishing protocol of \Cref{sec:method}, and (iii) presenting the
parametric-coefficient mode within the same architecture and analysis as
the loss-weight mode. We do not claim novelty for parameter-conditioned
PINNs per se; the contribution is the unified treatment and the
training protocol.

\paragraph{Extrapolation in $\lambda$ vs.\ in physical parameters.}
In the loss-weight mode the network's role is closest to
\citet{dosovitskiy2020}; in the parametric-coefficient mode it is
closest to \citet{arthurs2021} and to FNO / DeepONet, but parameterised
over a low-dimensional scalar input rather than a function-valued one.
The two modes share an architecture; the choice of $\lambda$ is
task-driven.

\section{Method}
\label{sec:method}

\subsection{Loss-conditional formulation}

A standard PINN \citep{raissi2019} approximates the solution of a
single PDE $F u = 0$ with auxiliary boundary, initial, and
(optionally) data constraints
$\{B u = g_B,\, I u = g_I,\, D u = g_D\}$ by minimizing the weighted
composite loss
\begin{equation}
\begin{aligned}
   \mathcal{L}^{\mathrm{PINN}}(\theta;\, w)
   &=
     w_F \,\| F u_\theta \|_{L^2}^2
     + w_B \,\| B u_\theta - g_B \|_{L^2}^2 \\
   &\quad
     + w_I \,\| I u_\theta - g_I \|_{L^2}^2
     + w_D \,\| D u_\theta - g_D \|_{L^2}^2 ,
\end{aligned}
\label{eq:standard-pinn}
\end{equation}
with respect to the network parameters $\theta$. The weight vector
$w = (w_F, w_B, w_I, w_D)$ determines the tradeoff between satisfying
the PDE residual and enforcing the auxiliary constraints. In practice,
these weights are typically selected by grid search or adaptive
reweighting strategies such as SA-PINN \citep{mcclenny2020},
ReLoBRaLo \citep{bischof2021}, or Causal-PINN
\citep{wang2024}.

LC-PINN replaces this fixed-weight optimization problem with a
continuous family of problems indexed by a conditioning variable
$\lambda \in \Lambda$, where
$\Lambda \subseteq \mathbb{R}^{d_\lambda}$ denotes either
(i) the space of admissible loss weights themselves
($\lambda = w$, $d_\lambda = 4$ for Burgers and Buckley--Leverett), or
(ii) a physical PDE parameter such as a wavenumber or mobility ratio
($\lambda = k$, $d_\lambda = 1$ for Helmholtz). The resulting
loss-conditional PINN (LC-PINN) is a single network
\begin{equation}
   u_\theta : \Omega \times \Lambda \to \mathbb{R},
   \qquad
   (x,\lambda) \mapsto u_\theta(x,\lambda),
\label{eq:lc-network}
\end{equation}
trained to minimize the expected loss over the parameter family:
\begin{equation}
   J(\theta)
   =
   \mathbb{E}_{\lambda \sim p_\lambda}
   \left[
       \mathcal{L}\bigl(
           \lambda;\,
           u_\theta(\cdot,\lambda)
       \bigr)
   \right],
\label{eq:lc-objective}
\end{equation}
where $p_\lambda$ is a fixed sampling distribution over $\Lambda$
(uniform unless stated otherwise).

Architecturally, the modification is minimal: the network receives an
additional input of dimension $d_\lambda$. Computationally, each
optimization step requires only a standard forward/backward pass at a
freshly sampled $\lambda$. Conceptually, however, the formulation
changes the role of loss balancing in PINNs. Instead of identifying a
single optimal weighting configuration, the network is trained across
the entire loss-weight space, learning a continuous family of
solutions indexed by $\lambda$. This extends the loss-conditional
networks of \citet{dosovitskiy2020} from generative loss-mixture
modeling to the residual-loss setting of physics-informed learning.

\subsection{Inference: $K$-shot averaging across $\lambda$}
\label{sec:k-shot}

At evaluation time we report a single rel-$L^2$ error per task. In the
loss-weight regime there is no canonical ``true'' choice of
$\lambda$, so following \citet{dosovitskiy2020} we average predictions
over $K$ independent samples drawn from $p_\lambda$:
\begin{equation}
   \hat u(x)
   =
   \frac{1}{K}
   \sum_{i=1}^{K}
   u_\theta(x,\lambda^{(i)}),
   \qquad
   \lambda^{(i)}
   \stackrel{\text{i.i.d.}}{\sim}
   p_\lambda .
\label{eq:k-shot}
\end{equation}

This $K$-shot averaging exposes the variability of LC-PINN across the
loss-weight family and highlights the amortization advantage of the
approach. A conventional PINN must be retrained $K$ times to generate
predictions for $K$ different weight configurations, whereas LC-PINN
produces all $K$ predictions from a single training run.

\paragraph{Reconciling $K$-shot averaging with $\lambda$-invariance.}
At first sight \Cref{prop:lc-invariance} below---which states that the
LC optimum is $\lambda$-invariant---seems to make $K$-shot averaging
either redundant or contradictory: if every $\lambda$-slice is the same
solution, why average over $\lambda$? The resolution is that invariance
is an asymptotic, almost-everywhere statement about the \emph{exact}
optimum, which a finite-capacity trained network realises only
approximately. We measure the residual variability directly. Drawing
$K=200$ weightings $\lambda^{(i)}\sim\mathcal{U}(0,1)^4$ and evaluating
$u_\theta(x,\lambda^{(i)})$ on the Burgers grid, the per-point standard
deviation across $i$ is $1.4\times10^{-3}$---about $0.19\%$ of the mean
solution magnitude. The network is therefore \emph{neither} collapsed to
a single effective $\lambda$ (the spread is nonzero) \emph{nor}
traversing a genuine trade-off (the spread is two-to-three orders of
magnitude below the signal). This is the key distinction from the
generative setting of \citet{dosovitskiy2020}, where the loss terms
genuinely conflict and conditioning traverses a Pareto surface: the
PINN residual, boundary, initial, and data losses share a common zero,
so the conditional family is nearly degenerate. $K$-shot averaging is
thus a mild variance-reduction step over an almost-invariant family
rather than a selection among competing solutions, which is exactly why
the reported rel-$L^2$ is flat in $K$ (the $K$-shot mean reaches its
$K\!\to\!\infty$ limit by $K\approx 25$).

\subsection{$\lambda$-invariance at the LC optimum}
\label{sec:theorem}

The central hypothesis behind LC-PINN is that a single conditional
network $u_\theta(x,\lambda)$, trained with
$\lambda \sim p_\lambda$, can simultaneously parameterize an entire
family of PDE solutions
$\{F_\lambda u = 0\}_{\lambda \in \Lambda}$. The following proposition
formalizes this statement at a global optimum of the LC-PINN objective.

\paragraph{Setup.}
Let $\Lambda \subseteq \mathbb{R}^{d_\lambda}$ denote the parameter
space, $p_\lambda$ a probability measure on $\Lambda$ with full
support, and $\Omega \subseteq \mathbb{R}^{d_x}$ the
spatial--temporal domain equipped with sampling measure $p_\Omega$.
Each $\lambda \in \Lambda$ defines a residual operator $F_\lambda$
together with associated boundary, initial, and optional data-fit
operators
$\{B_\lambda, I_\lambda, D_\lambda\}$. For a candidate field
$v : \Omega \to \mathbb{R}$, define the corresponding per-parameter
loss functional
\[
\begin{aligned}
   \mathcal{L}(\lambda; v)
   &=
     w_F \|F_\lambda v\|_{L^2(\Omega,p_\Omega)}^2
     + w_B \|B_\lambda v\|_{L^2(\partial\Omega)}^2 \\
   &\quad
     + w_I \|I_\lambda v\|_{L^2(t=0)}^2
     + w_D \|D_\lambda v\|_{L^2}^2 ,
\end{aligned}
\]
where $(w_F,w_B,w_I,w_D)$ are strictly positive coefficients. The
LC-PINN objective is the expected loss over the parameter family:
\[
   J(\theta)
   =
   \int_\Lambda
      \mathcal{L}\bigl(
          \lambda;\,
          u_\theta(\cdot,\lambda)
      \bigr)
   \,dp_\lambda(\lambda).
\]

We make the following assumptions:
\begin{itemize}\setlength{\itemsep}{0pt}
   \item[(A1)] For every $\lambda \in \Lambda$, the network class
         contains a residual minimizer
         $u^\star_\lambda \in \arg\min_v \mathcal{L}(\lambda;v)$ such
         that
         $\mathcal{L}(\lambda;u^\star_\lambda)=0$.

   \item[(A2)] The map
         $\lambda \mapsto
         \mathcal{L}(\lambda;u_\theta(\cdot,\lambda))$
         is $dp_\lambda$-integrable for every $\theta$.

   \item[(A3)] The network $u_\theta(x,\lambda)$ is jointly continuous
         in $(x,\lambda)$.
\end{itemize}

\begin{proposition}[Residual minimizers at the LC optimum]
\label{prop:lc-invariance}

Let $\theta^\star \in \arg\min_\theta J(\theta)$. Under assumptions
\textnormal{(A1)--(A3)}, the section
$u_{\theta^\star}(\cdot,\lambda)$ is a residual minimizer for
$p_\lambda$-almost every $\lambda \in \Lambda$:
\[
   \mathcal{L}\bigl(
      \lambda;\,
      u_{\theta^\star}(\cdot,\lambda)
   \bigr)
   =
   0,
   \qquad
   p_\lambda\text{-a.e. } \lambda \in \Lambda .
\]

In particular, on $\operatorname{supp}(p_\lambda)$ the LC-PINN
satisfies the PDE residual
$F_\lambda u_{\theta^\star}(\cdot,\lambda)=0$ in
$L^2(\Omega,p_\Omega)$ together with the associated auxiliary
constraints.
\end{proposition}

\paragraph{Sketch proof.}
For every $\theta$ and $\lambda$, the quantity
$\mathcal{L}(\lambda;u_\theta(\cdot,\lambda))$ is nonnegative since
each term is a squared $L^2$ norm. By assumption (A1), the minimum
per-parameter loss equals zero, implying that
$\inf_\theta J(\theta)\geq0$ and that this lower bound is attainable
whenever the network class can jointly represent the family
$\{u^\star_\lambda\}_{\lambda\in\Lambda}$.

Under assumption (A3), the universal approximation theorem for
$\tanh$ networks \citep{hornik1991,pinkus1999} guarantees that, on the
compact domain $\Omega \times \Lambda$, there exists
$\theta^\star \in \Theta$ such that
\[
   \sup_{(x,\lambda)\in\Omega\times\Lambda}
   |u_{\theta^\star}(x,\lambda)-u^\star_\lambda(x)|
\]
is arbitrarily small, and therefore $J(\theta^\star)$ can be made
arbitrarily close to zero. Consequently,
\[
   0
   =
   J(\theta^\star)
   =
   \int_\Lambda
      \mathcal{L}\bigl(
         \lambda;\,
         u_{\theta^\star}(\cdot,\lambda)
      \bigr)
   \,dp_\lambda(\lambda).
\]

Since the integrand is nonnegative, the standard vanishing-integral
argument implies
\[
   \mathcal{L}\bigl(
      \lambda;\,
      u_{\theta^\star}(\cdot,\lambda)
   \bigr)
   =
   0
\]
for $p_\lambda$-almost every $\lambda \in \Lambda$.
\hfill $\square$

\paragraph{Remark 1 (role of the support).}
The proposition is a $p_\lambda$-almost-everywhere statement.
Consequently, LC-PINN may behave arbitrarily on $p_\lambda$-null sets,
including boundary regions of $\Lambda$ that receive no probability
mass. In our experiments
$p_\lambda=\mathrm{Uniform}(\Lambda)$ has full support, so the
$p_\lambda$-a.e.\ statement coincides with Lebesgue-a.e.\ on
$\Lambda$.

\paragraph{Remark 2 (sampling-law invariance).}
If two probability measures $p_\lambda$ and $\tilde p_\lambda$ share
the same support $\Lambda_0 \subseteq \Lambda$, then their sets of
a.e.-residual minimizers on $\Lambda_0$ coincide. Thus, the asymptotic
optimum set of the LC objective depends only on the support of the
sampling distribution, not on its precise density. At finite training
budget, however, the choice of $p_\lambda$ affects how gradient signal
is distributed across parameter space, which we observe empirically in
\Cref{sec:results-ablations}.

\paragraph{Remark 3 (Monte Carlo estimation).}
\label{rem:sample}

In practice, each optimization step estimates
$\nabla_\theta J(\theta)$ using $n_\lambda$ i.i.d.\ samples drawn from
$p_\lambda$. The resulting Monte Carlo estimator is unbiased and has
variance $O(1/n_\lambda)$. Empirically,
$n_\lambda=4$ provides a good tradeoff between gradient variance
(small $n_\lambda$) and optimization progress per unit compute
(large $n_\lambda$), see \Cref{sec:results-ablations}.

\subsection{Network and training details}
\label{sec:training-details}

All experiments use fully-connected $\tanh$ networks with hidden
widths $[64,64,64,64]$, Xavier initialization, and a single
real-valued output head. The conditioning mechanism depends on the
problem regime.

In the \emph{loss-weight regime} (Burgers and Buckley--Leverett),
the conditioning variable $\lambda$ is concatenated directly to the
spatial--temporal coordinates at the input layer, following the
loss-conditional formulation of \citet{dosovitskiy2020}. In the
\emph{parametric-coefficient regime} (1D/2D/3D Helmholtz and
Schrödinger equations), $\lambda$ is instead mapped to FiLM
\citep{perez2018film} scale-and-shift parameters that modulate each
hidden layer. Training is then followed by an L-BFGS refinement phase
equipped with a revert-on-worse safeguard.

This regime-dependent choice is itself one of our main empirical
findings, and it is not obvious a priori. \emph{FiLM helps when
$\lambda$ changes the PDE operator but hurts when $\lambda$ only
re-weights existing losses.} Concretely, in the parametric-coefficient
regime the FiLM+L-BFGS combination reduces rel-$L^2$ by approximately an
order of magnitude over simple input concatenation, whereas in the
loss-weight regime the opposite holds and plain concatenation is more
stable (full ablation in \Cref{sec:appendix-film-loss-weight}). This is
precisely why we present the two regimes together rather than as
separate methods: they share one architecture and one analysis, and the
controlled concat-vs-FiLM ablation is what distinguishes them. The
practical rule of thumb---multiplicative (FiLM) conditioning for
operator-changing parameters, additive (concat) conditioning for
loss-reweighting parameters---transfers directly to other
conditional-PINN settings.

Optimization uses Adam with learning rate
$\eta = 10^{-3}$, cosine learning-rate annealing, and gradient-norm
clipping at $1.0$. At each optimization step, parameters
$\lambda \sim p_\lambda$ are sampled independently and the
corresponding per-instance residual loss is evaluated on the current
batch of collocation points. Before entering the network,
$\lambda$ is normalized to the interval $[-1,1]$.

All experiments are implemented in PyTorch~2.9 and executed on an
Apple~M4 Max GPU through the MPS backend. Full hyperparameter settings
and collocation budgets for each PDE family are reported in
\Cref{sec:appendix-hyper}.

\section{Local effect of loss-conditioning on PINN optimization}
\label{sec:local_lc_pinn_theory}

PINNs are typically trained by minimizing a weighted combination of
multiple nonnegative objectives, including PDE residual, boundary,
initial-condition, and data losses:
\[
    \mathcal{L}_{\lambda}(\theta)
    =
    \sum_{i=1}^m
    \lambda_i \mathcal{L}_i(\theta),
    \qquad
    \lambda_i \geq 0 .
\]

In the ideal well-specified setting, these objectives are compatible:
there exists a parameter vector $\theta^\star$ such that
\[
    \mathcal{L}_1(\theta^\star)
    =
    \cdots
    =
    \mathcal{L}_m(\theta^\star)
    =
    0 .
\]
Consequently, loss weighting does not change the target solution
itself, but rather modifies the local optimization geometry around the
shared optimum.

To analyze this effect, we consider a local quadratic approximation in
a neighbourhood of $\theta^\star$. Assume that each loss component
admits the second-order expansion
\[
    \mathcal{L}_i(\theta)
    =
    \frac12
    (\theta-\theta^\star)^\top
    H_i
    (\theta-\theta^\star),
    \qquad
    H_i \succeq 0,
    \quad
    i=1,\dots,m .
\]
For a fixed loss-weight vector $\lambda$, the resulting weighted PINN
objective becomes
\[
    \mathcal{L}_{\lambda}(\theta)
    =
    \frac12
    (\theta-\theta^\star)^\top
    H_\lambda
    (\theta-\theta^\star),
\]
where
\[
    H_\lambda
    =
    \sum_{i=1}^m
    \lambda_i H_i .
\]

The matrix $H_\lambda$ determines the local conditioning of the
optimization problem. In particular, small eigenvalues of
$H_\lambda$ correspond to weakly controlled directions in parameter
space and may substantially slow gradient-based optimization.

We now compare this classical parameterization with the
loss-conditional one. Let $\theta$ denote the original PINN
parameters and let $\eta$ denote the additional trainable parameters
introduced by conditioning the network on the loss-weight vector
$\lambda$. For fixed $\lambda$, the conditional network can be viewed
locally as a standard PINN with effective parameters
\[
    \theta_{\mathrm{eff}}(\lambda;\theta,\eta)
    =
    \theta + B(\lambda,\eta),
\]
where $B$ is bilinear in $\lambda$ and $\eta$. Equivalently, there
exist matrices $M_1,\dots,M_m$ such that
\[
    B(\lambda,\eta)
    =
    \sum_{j=1}^m
    \lambda_j M_j \eta .
\]

The following theorem shows that, at the same effective parameter
vector, the loss-conditional parameterization possesses an
instantaneous loss-decay rate at least as large as that of the
standard PINN.

\begin{theorem}[Local loss-decay advantage of loss-conditioning]
\label{thm:local_decay_lc_pinn}

Fix $\lambda$ and consider the local quadratic objective
\[
    \mathcal{L}_{\lambda}(\vartheta)
    =
    \frac12
    (\vartheta-\theta^\star)^\top
    H_\lambda
    (\vartheta-\theta^\star),
    \qquad
    H_\lambda \succeq 0 .
\]

Consider two parameterizations of the same effective parameter vector
$\vartheta$:
\[
    \vartheta = \theta_{\mathrm{reg}}
\]
for the standard PINN, and
\[
    \vartheta
    =
    \theta_{\mathrm{eff}}(\lambda;\theta,\eta)
    =
    \theta + B(\lambda,\eta)
\]
for the loss-conditional PINN.

Suppose that, at some time instant, both parameterizations represent
the same effective point:
\[
    \theta_{\mathrm{reg}}
    =
    \theta_{\mathrm{eff}}(\lambda;\theta,\eta)
    =
    \vartheta .
\]

Then, under continuous-time gradient descent, the instantaneous decay
rate of the loss in the loss-conditional parameterization satisfies
\[
\begin{aligned}
    -\frac{d}{dt}\,
    \mathcal{L}_{\lambda}
    \!\bigl(
        \theta_{\mathrm{eff}}(\lambda;\theta,\eta)
    \bigr)
    =\;&
    -\frac{d}{dt}\,
    \mathcal{L}_{\lambda}
    (\theta_{\mathrm{reg}}) \\
    &+
    \left\|
        D_\eta B(\lambda,\eta)^\top
        \nabla_\vartheta
        \mathcal{L}_{\lambda}(\vartheta)
    \right\|^2 .
\end{aligned}
\]

In particular,
\[
    -\frac{d}{dt}
    \mathcal{L}_{\lambda}
    \bigl(
        \theta_{\mathrm{eff}}(\lambda;\theta,\eta)
    \bigr)
    \geq
    -\frac{d}{dt}
    \mathcal{L}_{\lambda}
    (\theta_{\mathrm{reg}}).
\]

If
\[
    B(\lambda,\eta)
    =
    \sum_{j=1}^m
    \lambda_j M_j \eta,
\]
then
\[
    D_\eta B(\lambda,\eta)
    =
    \sum_{j=1}^m
    \lambda_j M_j,
\]
and the additional decay term becomes
\[
    \left\|
        \left(
            \sum_{j=1}^m
            \lambda_j M_j
        \right)^\top
        H_\lambda
        (\vartheta-\theta^\star)
    \right\|^2 .
\]
\end{theorem}

The theorem provides a local explanation for the empirical robustness
of LC-PINN. Loss-conditioning does not alter the common optimum
$\theta^\star$; instead, it enriches the parameterization through
additional conditional directions in parameter space. Whenever the
regular and conditional models represent the same effective PINN, the
conditional model retains the original descent directions while also
acquiring additional optimization directions through the conditioning
parameters.

Consequently, in the local quadratic regime, the instantaneous loss
decrease of the conditional model exceeds that of the standard PINN by
a nonnegative squared-gradient term. Intuitively, this improvement
arises because the optimization is performed in a higher-dimensional
parameter space with additional degrees of freedom, allowing the model
to identify more effective descent directions during training.

\paragraph{Scope of this result.}
We state plainly what \Cref{thm:local_decay_lc_pinn} does and does not
establish. The extra descent term is a generic consequence of
optimising in an enlarged parameter space and is not specific to
loss-conditioning: any added, non-degenerate parameterisation would
yield an analogous instantaneous advantage. The result is local (a
quadratic model) and instantaneous (one gradient step), so it is best
read as a consistency check---loss-conditioning does not \emph{harm}
the local decay rate---rather than as an explanation of the
order-of-magnitude accuracy gains we observe empirically. Those gains
are cumulative and stem from the smooth-in-$\lambda$ parameter sharing
that the conditional family induces over the whole trajectory, a
mechanism this local analysis does not capture and which we do not
claim to prove here.

\section{Results}
\label{sec:results}

\begin{figure}[!t]
\centering
\includegraphics[width=\linewidth]{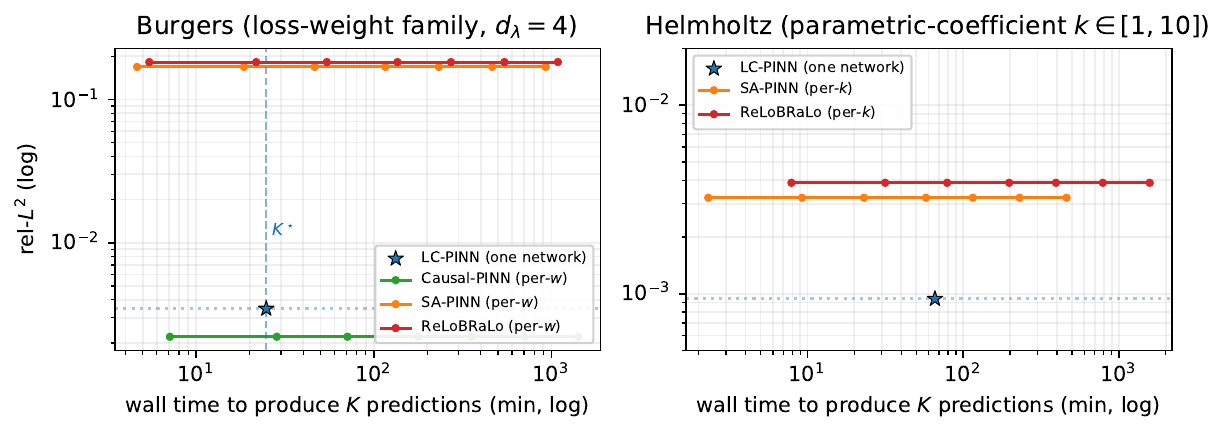}
\caption{Wall-time/accuracy frontier as the number $K$ of evaluated
$\lambda$ values increases. LC-PINN trains once and produces
predictions for any $K$, while per-$\lambda$ baselines move rightward
by a factor of $K$. On Helmholtz, LC-PINN Pareto-dominates from
$K=1$. On Burgers, Causal-PINN is more accurate for a single
configuration, but becomes more expensive than LC-PINN once
$K \gtrsim 4$.}
\label{fig:pareto}
\end{figure}

We evaluate LC-PINN in two amortisation regimes. The first is the
parametric-coefficient regime, where $\lambda$ is a physical PDE
parameter and the comparison with operator-learning methods is most
direct. The second is the loss-weight regime, where $\lambda$ is a
loss-weight vector and LC-PINN is compared with adaptive PINN
balancers. For the parametric-coefficient experiments, we report
rel-$L^2$ error averaged over a fixed parameter grid. For the
loss-weight experiments, we report $K$-shot averaged rel-$L^2$ error.
Unless stated otherwise, all results are mean $\pm$ standard deviation
over four random seeds; per-parameter baselines use two seeds when
their retraining cost is substantially higher. \Cref{fig:pareto}
summarizes the main wall-time/accuracy tradeoff.

\subsection{1D Helmholtz: parametric-coefficient regime}
\label{sec:results-helmholtz}

\Cref{tab:helmholtz} reports rel-$L^2$ error averaged over the
five-point grid
$k \in \{1.00,3.25,5.50,7.75,10.00\}$. LC-PINN receives the
wavenumber $k$ as a normalized input, whereas the baselines are
retrained separately for each value of $k$.

\begin{table}[!t]
\centering
\caption{1D Helmholtz, rel-$L^2$ averaged across
$k \in \{1.00, 3.25, 5.50, 7.75, 10.00\}$. LC-PINN is one trained
network evaluated at each $k$; baselines are five separate
retrainings, one per $k$.}
\label{tab:helmholtz}
\small
\begin{tabular}{lcc}
\toprule
Method & rel-$L^2$ (mean $\pm$ std) & wall time \\
\midrule
LC-PINN (FiLM+L-BFGS, one network) & $\mathbf{9.39 \times 10^{-4} \pm 1.18 \times 10^{-4}}$ & 65.9\,min/seed \\
PI-DeepONet (one network)          & $2.21 \times 10^{-2} \pm 3.18 \times 10^{-2}$ & 61.8\,min/seed \\
SA-PINN (per-$k$, 5 retrainings)   & $3.23 \times 10^{-3} \pm 2.90 \times 10^{-3}$ & 11.4\,min/$k$ \\
ReLoBRaLo (per-$k$, 5 retrainings) & $3.87 \times 10^{-3} \pm 2.82 \times 10^{-3}$ & 7.9\,min/$k$ \\
\bottomrule
\end{tabular}
\end{table}

With FiLM conditioning on $k$ and an L-BFGS refinement phase, a single
LC-PINN covers the full interval $k \in [1,10]$ with average
rel-$L^2 \approx 9 \times 10^{-4}$. This improves over SA-PINN by
approximately $3.4\times$ and over ReLoBRaLo by approximately
$4.1\times$, while requiring less total wall time than running all
five per-$k$ retrainings.

\paragraph{Per-parameter breakdown.}
The grid average hides an important difference between the methods.
\Cref{tab:helm-per-k} reports rel-$L^2$ error at each value of $k$.
Retrained baselines can be extremely accurate at easy parameter values:
for example, ReLoBRaLo reaches $5.75 \times 10^{-6}$ at $k=1$.
However, their performance varies sharply across the family. At
$k=10$, ReLoBRaLo deteriorates to $1.57 \times 10^{-2}$, three orders
of magnitude worse than its own $k=1$ result. SA-PINN exhibits similar
instability, with its worst result occurring at $k=3.25$.

By contrast, LC-PINN remains within one order of magnitude of its best
performance across the entire grid. This highlights the main empirical
tradeoff: amortisation may sacrifice peak accuracy at the easiest
single parameter values, but it provides substantially greater
coherence across the full parameter family. LC-PINN is therefore aimed
at parametric settings where solutions are needed for many values of
$k$, rather than at single-parameter optimization.

\begin{table}[!t]
\centering
\caption{Per-$k$ rel-$L^2$ on 1D parametric Helmholtz
(mean $\pm$ std over four seeds). Bold indicates the best method per
row. LC-PINN is one trained network evaluated at each $k$; baselines
are retrained separately for each $k$.}
\label{tab:helm-per-k}
\small
\begin{tabular}{r|ccc}
\toprule
$k$ & SA-PINN (per-$k$) & ReLoBRaLo (per-$k$) & LC-PINN (one net) \\
\midrule
$1.00$  & $(1.23\!\pm\!0.61)\!\times\!10^{-4}$  & $\mathbf{(5.75\!\pm\!1.73)\!\times\!10^{-6}}$ & $(1.25\!\pm\!0.37)\!\times\!10^{-3}$ \\
$3.25$  & $(1.29\!\pm\!1.23)\!\times\!10^{-2}$  & $\mathbf{(7.04\!\pm\!2.31)\!\times\!10^{-5}}$ & $(1.27\!\pm\!0.67)\!\times\!10^{-3}$ \\
$5.50$  & $(4.83\!\pm\!4.75)\!\times\!10^{-4}$  & $\mathbf{(1.29\!\pm\!0.30)\!\times\!10^{-4}}$ & $(7.97\!\pm\!8.53)\!\times\!10^{-4}$ \\
$7.75$  & $(1.14\!\pm\!1.16)\!\times\!10^{-3}$  & $(3.42\!\pm\!1.04)\!\times\!10^{-3}$ & $\mathbf{(2.64\!\pm\!0.82)\!\times\!10^{-4}}$ \\
$10.00$ & $(1.50\!\pm\!1.68)\!\times\!10^{-3}$  & $(1.57\!\pm\!1.13)\!\times\!10^{-2}$ & $\mathbf{(1.11\!\pm\!0.86)\!\times\!10^{-3}}$ \\
\midrule
\textbf{grid mean} & $3.23\!\times\!10^{-3}$ & $3.87\!\times\!10^{-3}$ & $\mathbf{9.39\!\times\!10^{-4}}$ \\
\bottomrule
\end{tabular}
\end{table}

\begin{figure}[!t]
\centering
\includegraphics[width=0.6\linewidth]{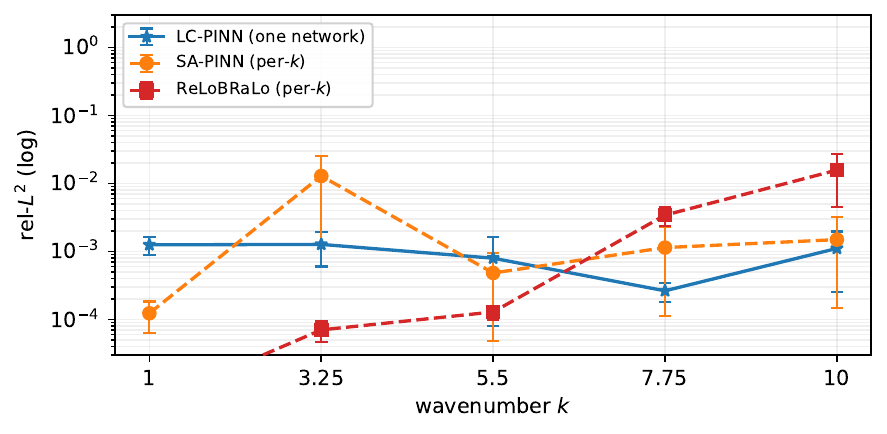}
\caption{Per-$k$ rel-$L^2$ on 1D parametric Helmholtz. Error bars
show $\pm$ standard deviation over four seeds. Retrained baselines
vary strongly across the parameter grid, whereas LC-PINN remains
comparatively stable.}
\label{fig:per_k_helmholtz}
\end{figure}

Overall, the Helmholtz experiment shows that LC-PINN improves the
metric that matters in the parametric setting: grid-averaged and
worst-case accuracy across the family.

\subsection{1D Schrödinger: parametric-coefficient regime}
\label{sec:results-schrodinger}

We next test whether the improvement in \Cref{sec:results-helmholtz}
is specific to the Helmholtz operator. We consider a 1D stationary
Schrödinger problem with parametric harmonic potential,
\[
    -u'' + \alpha^2(x-\tfrac12)^2 u = f(x;\alpha),
    \qquad x \in (0,1),
\]
with homogeneous Dirichlet boundary conditions and
$\alpha \in [0.5,10]$. Compared with Helmholtz, the parameter
$\alpha$ multiplies a spatially varying potential rather than a
constant zeroth-order term. The manufactured solution is
$\sin(\pi x)\exp(-\alpha(x-\tfrac12)^2/2)$; details of the forcing are
given in \Cref{sec:appendix-hyper}.

\begin{table}[!t]
\centering
\caption{1D Schrödinger equation with parametric harmonic well,
rel-$L^2$ averaged across $\alpha \in \{0.5,5.0,10.0\}$. LC-PINN is
one trained network evaluated at each $\alpha$; SA-PINN is retrained
separately for each $\alpha$.}
\label{tab:schrodinger}
\small
\begin{tabular}{lcc}
\toprule
Method & rel-$L^2$ (mean $\pm$ std) & wall time \\
\midrule
LC-PINN (FiLM+L-BFGS, one network)        & $\mathbf{1.99 \times 10^{-4} \pm 1.02 \times 10^{-4}}$ & 46.6\,min/seed \\
PI-DeepONet (one network)                 & $1.79 \times 10^{-4} \pm 3.70 \times 10^{-5}$        & 40.3\,min/seed \\
SA-PINN (per-$\alpha$, 3 retrainings)     & $3.89 \times 10^{-3} \pm 3.70 \times 10^{-3}$ & 0.8\,min/$\alpha$ \\
\bottomrule
\end{tabular}
\end{table}

LC-PINN improves over the per-$\alpha$ SA-PINN baseline by roughly
$20\times$ on the grid mean. The per-$\alpha$ breakdown
(\Cref{tab:schrod-per-alpha}) again shows that retrained baselines may
perform well at individual parameter values but can be unstable across
the family: at $\alpha=5.0$, the two SA-PINN seeds differ by two
orders of magnitude. LC-PINN maintains seed-to-seed standard deviation
below $3.2\times10^{-4}$ throughout $\alpha \in [0.5,10]$.

PI-DeepONet matches LC-PINN on this smoother parametric potential,
whereas LC-PINN substantially outperforms PI-DeepONet on Helmholtz.
This suggests that amortisation is robust across architectures for
smooth potentials, while high-frequency wave operators particularly
benefit from FiLM-based conditioning.

\subsection{2D Helmholtz: parametric-coefficient regime}
\label{sec:results-helmholtz-2d}

We extend the Helmholtz experiment to the unit square:
\[
    \Delta u + k^2 u = f(x,y;k),
    \qquad (x,y) \in [0,1]^2,
\]
with homogeneous Dirichlet boundary conditions and manufactured
solution
\[
    u(x,y;k)
    =
    \sin(\pi x)\sin(\pi y)\cos(kx)\cos(ky).
\]
The same LC-PINN architecture conditions on the normalized wavenumber
$k$, while the baselines are retrained separately for each training
value of $k$.

\begin{table}[!t]
\centering
\caption{2D Helmholtz on the unit square, rel-$L^2$ averaged across
$k \in \{1.00,5.50,10.00\}$. LC-PINN is one trained network evaluated
at each $k$; baselines are retrained separately for each $k$.}
\label{tab:helmholtz-2d}
\small
\begin{tabular}{lcc}
\toprule
Method & rel-$L^2$ (mean $\pm$ std) & wall time \\
\midrule
LC-PINN (FiLM+L-BFGS, one network) & $\mathbf{2.36 \times 10^{-2} \pm 2.56 \times 10^{-2}}$ & 85.9\,min/seed \\
SA-PINN (per-$k$, 3 retrainings)   & $7.83 \times 10^{-2} \pm 2.76 \times 10^{-2}$ & 3.0\,min/$k$ \\
ReLoBRaLo (per-$k$, 3 retrainings) & $1.13 \times 10^{-1} \pm 1.72 \times 10^{-2}$ & 5.6\,min/$k$ \\
\bottomrule
\end{tabular}
\end{table}

The 2D problem has a richer solution structure, with boundary nodal
lines and tensor-product oscillations in both spatial directions.
Nevertheless, the same LC-PINN construction transfers without
modification. It reaches average rel-$L^2 \approx 2.4\times10^{-2}$,
outperforming SA-PINN by approximately $3.3\times$ and ReLoBRaLo by
approximately $4.8\times$.

\subsection{3D Helmholtz: parametric-coefficient regime}
\label{sec:results-helmholtz-3d}

Finally, we consider 3D Helmholtz on the unit cube,
\[
    \Delta u + k^2 u = f(x,y,z;k),
    \qquad (x,y,z) \in [0,1]^3,
\]
with homogeneous Dirichlet boundary conditions and tensor-product
manufactured solution
\[
    u(x,y,z;k)
    =
    \prod_{q\in\{x,y,z\}}
    \sin(\pi q)\cos(kq).
\]
The evaluation grid is $k \in \{1,3,5\}$; the high-$k$ regime becomes
substantially harder in 3D because oscillatory complexity increases in
all spatial directions.

\begin{table}[!t]
\centering
\caption{3D Helmholtz on the unit cube, rel-$L^2$ averaged across
$k \in \{1,3,5\}$. LC-PINN is one trained network evaluated at each
$k$; baselines are retrained separately for each $k$.}
\label{tab:helmholtz-3d}
\small
\begin{tabular}{lcc}
\toprule
Method & rel-$L^2$ (mean $\pm$ std) & wall time \\
\midrule
LC-PINN (FiLM+L-BFGS, one network) & $\mathbf{1.67 \times 10^{-2} \pm 1.97 \times 10^{-3}}$ & 228.9\,min/seed \\
SA-PINN (per-$k$, 3 retrainings)   & $6.97 \times 10^{-2} \pm 2.47 \times 10^{-3}$ & 5.7\,min/$k$ \\
ReLoBRaLo (per-$k$, 3 retrainings) & $1.03 \times 10^{-1} \pm 7.82 \times 10^{-3}$ & 11.9\,min/$k$ \\
\bottomrule
\end{tabular}
\end{table}

LC-PINN again remains the strongest method on the grid mean, improving
over SA-PINN by approximately $4.2\times$ and over ReLoBRaLo by
approximately $6.2\times$. This confirms that the parametric
conditioning approach extends from 1D to 2D and 3D without changing
the core method.

\subsection{Loss-weight regime: Burgers and Buckley--Leverett}
\label{sec:results-burgers-bl}

We now turn to loss-weight conditioning, where
$\lambda \in \Delta^3$ parameterizes the relative weights of the PINN
loss components. In this regime, FiLM modulation and L-BFGS refinement
do not improve stability; we therefore use the simpler
concat+Adam configuration following \citet{dosovitskiy2020}. 
\Cref{tab:burgers} reports rel-$L^2$ error on viscous Burgers with
$\nu=0.01/\pi$, initial condition $u(0,x)=-\sin(\pi x)$, and final
time $t=1$.

\begin{table}[!t]
\centering
\caption{Burgers, rel-$L^2$ at the test initial condition
$-\sin(\pi x)$. LC-PINN reports $K=100$-shot averaging across
loss weights; baselines are single-shot at their learned or
self-adapted weighting.}
\label{tab:burgers}
\small
\begin{tabular}{lcc}
\toprule
Method & rel-$L^2$ (mean $\pm$ std) & wall time / seed \\
\midrule
LC-PINN ($K\!=\!100$) & $3.47 \times 10^{-3} \pm 2.73 \times 10^{-3}$ & 24.8\,min \\
Causal-PINN           & $2.21 \times 10^{-3} \pm 6.72 \times 10^{-4}$ &  7.1\,min \\
SA-PINN               & $1.68 \times 10^{-1} \pm 3.87 \times 10^{-2}$ &  4.6\,min \\
ReLoBRaLo             & $1.82 \times 10^{-1} \pm 1.71 \times 10^{-2}$ &  5.4\,min \\
FNO (operator)        & $5.37 \times 10^{-1} \pm 1.89 \times 10^{-1}$ &  8.0\,min \\
\bottomrule
\end{tabular}
\end{table}

LC-PINN matches the strongest single-shot baseline, Causal-PINN,
within a factor of two and outperforms SA-PINN and ReLoBRaLo by more
than an order of magnitude. Its main advantage is amortisation: once
trained, the same network produces predictions for arbitrary loss
weights. The amortisation factor
$A(K)=K\,T_{\mathrm{base}}/T_{\mathrm{LC}}$ breaks even at
$K^\star \approx 3.5$--$5.4$ and reaches roughly $19$--$29\times$ at
$K=100$.

On viscous Buckley--Leverett with $\varepsilon=10^{-2}$
(\Cref{tab:bl}), ReLoBRaLo is approximately $2\times$ more accurate
than LC-PINN at a single weighting, while SA-PINN fails to converge
below rel-$L^2 \approx 0.5$. In the zero-viscosity formulation, all
PINN-family methods saturate near this error level. We state this
plainly: on the smooth, viscous regime where surrogate methods actually
work, LC-PINN is competitive (within a factor of two) but not the best
single-shot method, and the strongest per-instance method---ReLoBRaLo
here, Causal-PINN on Burgers---remains a different one. We include
Buckley--Leverett not to claim a porous-media state of the art but as
evidence that the loss-conditional construction extends to that
application regime in principle; the contribution is amortisation across
the weight family, not per-weighting accuracy.

\begin{table}[!t]
\centering
\caption{Viscous-regularized Buckley--Leverett
($\varepsilon=10^{-2}$), rel-$L^2$ at the test snapshots. The
reference is produced by an in-house finite-volume solver and used
only for evaluation. LC-PINN reports $K=25$-shot averaging over
loss weights; baselines are single-shot.}
\label{tab:bl}
\small
\begin{tabular}{lc}
\toprule
Method & rel-$L^2$ (mean $\pm$ std) \\
\midrule
LC-PINN ($K\!=\!25$) & $1.03 \times 10^{-2} \pm 1.30 \times 10^{-3}$ \\
SA-PINN              & $4.60 \times 10^{-1} \pm 6.74 \times 10^{-2}$ \\
ReLoBRaLo            & $5.09 \times 10^{-3} \pm 7.79 \times 10^{-4}$ \\
\bottomrule
\end{tabular}
\end{table}

\subsection{Amortisation crossover across families}
\label{sec:results-amortisation}

The value of LC-PINN is amortisation, so we make the trade-off
quantitative for every family using the per-method wall times already
reported above (no new experiments). Let $T_{\mathrm{LC}}$ be the cost
of the single LC-PINN training and $T_{\mathrm{base}}$ the cost of
retraining the strongest per-instance baseline once. LC-PINN's one
network is cheaper than a sweep as soon as the number of instances
exceeds the break-even count $K^\star = T_{\mathrm{LC}}/T_{\mathrm{base}}$
(\Cref{tab:amortisation}).

\begin{table}[!t]
\centering
\caption{Amortisation break-even $K^\star$: the number of parameter
instances at which one LC-PINN training equals the cumulative cost of
retraining the strongest per-instance baseline. LC-PINN is favourable
above $K^\star$. It pays off quickly when per-instance retraining is
expensive (high-dimensional Helmholtz) and only for large sweeps when it
is cheap (Schrödinger).}
\label{tab:amortisation}
\small
\begin{tabular}{lccc}
\toprule
Family & $T_{\mathrm{LC}}$ & strongest baseline & $K^\star$ \\
\midrule
Burgers (loss-weight) & 24.8\,min & Causal-PINN 7.1\,min & $\approx 3.5$ \\
Helmholtz 1D & 65.9\,min & ReLoBRaLo 7.9\,min/$k$ & $\approx 8.3$ \\
Helmholtz 2D & 85.9\,min & ReLoBRaLo 5.6\,min/$k$ & $\approx 15$ \\
Helmholtz 3D & 228.9\,min & ReLoBRaLo 11.9\,min/$k$ & $\approx 19$ \\
Schrödinger & 46.6\,min & SA-PINN 0.8\,min/$\alpha$ & $\approx 58$ \\
\bottomrule
\end{tabular}
\end{table}

The picture is deliberately not uniformly favourable: where a single
retraining is cheap (Schrödinger, $0.8$\,min) the break-even is high and
LC-PINN only repays its cost for large sweeps, whereas in the regimes
where per-instance PINN training is genuinely expensive---the loss-weight
sweep and the higher-dimensional Helmholtz families---the crossover is
reached at a handful to a few tens of instances, well within a typical
parameter study.

\subsection{Ablations}
\label{sec:results-ablations}

We perform three ablations on Burgers; full results are reported in
\Cref{sec:appendix-ablations}. First, increasing the number of
evaluation samples $K$ has little effect: $K$-shot averaging remains
stable across $K\in\{25,\ldots,400\}$ at approximately
$3.5\times10^{-3}$. Second, the number of parameter samples per
optimization step is non-monotone, with the best performance at
$n_\lambda=4$. Third, replacing uniform sampling of $\lambda$ with a
log-uniform distribution degrades performance by approximately
$14\times$, highlighting the importance of finite-budget coverage of
the loss-weight space.

\section{Discussion}
\label{sec:discussion}

We introduced LC-PINN, a loss-conditional physics-informed neural network that treats loss weights and PDE parameters as conditioning variables rather than fixed hyperparameters. Instead of searching for a single optimal weighting configuration, LC-PINN learns a continuous family of solutions indexed by $\lambda$. This rethinks loss balancing in PINNs: rather than tuning one scalarization of a multi-objective objective, the network is trained across the entire loss-weight space.

Empirically, LC-PINN produces more coherent behaviour across parameter families than conventional per-configuration retraining. In the parametric-coefficient regime, FiLM-conditioned LC-PINNs achieve stable accuracy across Helmholtz and Schrödinger families while avoiding the strong variability observed in per-$k$ retrained baselines. In the loss-weight regime, a single LC-PINN approaches the performance of adaptive-weight PINNs while amortizing training across the full family of loss weights. The local analysis in Section~\ref{sec:local_lc_pinn_theory} suggests that conditioning improves optimization geometry by introducing additional descent directions in parameter space.

Conceptually, LC-PINN occupies an intermediate position between classical PINNs and operator-learning methods. Like neural operators, it amortizes computation across PDE families, but unlike operator-learning approaches it remains fully residual-based and does not require solver-generated paired datasets. The current formulation nevertheless becomes more difficult in regimes with shocks, bifurcations, or highly oscillatory solutions, where jointly approximating the full parameter family is challenging. Future work includes adaptive sampling strategies and extensions to inverse and multi-physics problems.

Overall, the results suggest that conditioning on loss geometry and PDE parameters can transform PINNs from single-instance solvers into amortized parametric surrogate models.

\bibliographystyle{unsrtnat}
\bibliography{bibliobase}

\newpage
\appendix
\section{Derivation of the local loss-decay result}
\label{app:local_decay_lc_pinn}

This appendix provides the details behind
Theorem~\ref{thm:local_decay_lc_pinn}. The goal is not to give a global
optimization guarantee, but to formalize a local mechanism by which
loss-conditioning can improve the instantaneous decay of a weighted PINN
loss near a common optimum.

\subsection{Local quadratic model for multi-objective PINN losses}

A PINN objective is typically a weighted combination of several
nonnegative loss components:
\[
    \mathcal{L}_{\lambda}(\theta)
    =
    \sum_{i=1}^m \lambda_i \mathcal{L}_i(\theta),
    \qquad
    \lambda_i \geq 0 .
\]
The components may correspond to the PDE residual, boundary conditions,
initial conditions, and data mismatch. In the well-specified case, there
exists a common solution $\theta^\star$ such that
\[
    \mathcal{L}_i(\theta^\star)=0,
    \qquad
    i=1,\dots,m .
\]
Therefore, all positive scalarizations share the same global zero-loss
solution.

Near $\theta^\star$, we approximate each component by its second-order
Taylor expansion. Since $\theta^\star$ is a local minimizer of each
nonnegative loss component, the first-order term vanishes. Thus,
locally,
\[
    \mathcal{L}_i(\theta)
    =
    \frac12
    (\theta-\theta^\star)^\top
    H_i
    (\theta-\theta^\star),
    \qquad
    H_i \succeq 0 .
\]
The weighted loss is then
\[
    \mathcal{L}_{\lambda}(\theta)
    =
    \sum_{i=1}^m \lambda_i \mathcal{L}_i(\theta)
    =
    \frac12
    (\theta-\theta^\star)^\top
    H_\lambda
    (\theta-\theta^\star),
\]
where
\[
    H_\lambda
    =
    \sum_{i=1}^m \lambda_i H_i .
\]
Thus, choosing the loss weights is equivalent, locally, to choosing the
effective Hessian $H_\lambda$ seen by the optimizer.

For the regular PINN parameterization, the gradient flow is
\[
    \dot{\theta}(t)
    =
    -
    \nabla_\theta
    \mathcal{L}_{\lambda}(\theta(t))
    =
    -
    H_\lambda
    (\theta(t)-\theta^\star).
\]
If $H_\lambda \succ 0$, then
\[
    \mathcal{L}_{\lambda}(\theta(t))
    \leq
    e^{-2\lambda_{\min}(H_\lambda)t}
    \mathcal{L}_{\lambda}(\theta(0)).
\]
Hence, the local convergence rate is controlled by the smallest
eigenvalue of $H_\lambda$. If $H_\lambda$ is nearly singular, convergence
can be slow in weakly controlled directions.

\subsection{Effective parameters induced by loss-conditioning}

We now describe the local parameterization induced by conditioning the
network on the loss-weight vector $\lambda$.

Let $\theta \in \mathbb{R}^p$ denote the parameters of the original PINN,
and let $\eta \in \mathbb{R}^q$ denote the additional trainable
parameters introduced by the conditional architecture. For example, if
the first layer of a standard PINN is
\[
    z_1 = W_x x + b,
\]
then concatenating $\lambda$ to the input gives
\[
    z_1 = W_x x + W_\lambda \lambda + b .
\]
For fixed $\lambda$, the term $W_\lambda \lambda$ acts as a
$\lambda$-dependent shift of the first-layer bias:
\[
    b_{\mathrm{eff}}(\lambda)
    =
    b + W_\lambda \lambda .
\]
Thus, after freezing $\lambda$, the conditional network can be viewed
locally as an ordinary PINN with $\lambda$-dependent effective
parameters.

More generally, we model this effect by
\[
    \theta_{\mathrm{eff}}(\lambda;\theta,\eta)
    =
    \theta + B(\lambda,\eta),
\]
where $B$ is bilinear in the conditioning variable $\lambda$ and in the
additional trainable parameters $\eta$. Equivalently, there exist
matrices
\[
    M_1,\dots,M_m \in \mathbb{R}^{p \times q}
\]
such that
\[
    B(\lambda,\eta)
    =
    \sum_{j=1}^m \lambda_j M_j \eta .
\]
The dependence on $\lambda$ is linear because the conditioning variables
enter the network input linearly, while the dependence on $\eta$ is
linear because $\eta$ represents the trainable weights connecting the
conditioning variables to the network.

Under this local model, the loss-conditional objective for fixed
$\lambda$ becomes
\[
    \mathcal{L}_{\lambda}
    \bigl(
        \theta_{\mathrm{eff}}(\lambda;\theta,\eta)
    \bigr)
    =
    \frac12
    \bigl(
        \theta + B(\lambda,\eta)-\theta^\star
    \bigr)^\top
    H_\lambda
    \bigl(
        \theta + B(\lambda,\eta)-\theta^\star
    \bigr).
\]
If the training samples $\lambda$ from a distribution $p(\lambda)$, the
corresponding expected local objective is
\[
    \mathcal{J}(\theta,\eta)
    =
    \mathbb{E}_{\lambda \sim p}
    \left[
        \frac12
        \bigl(
            \theta + B(\lambda,\eta)-\theta^\star
        \bigr)^\top
        H_\lambda
        \bigl(
            \theta + B(\lambda,\eta)-\theta^\star
        \bigr)
    \right].
\]
The theorem in the main text studies the instantaneous loss decay for a
fixed value of $\lambda$.

\subsection{Proof of Theorem~\ref{thm:local_decay_lc_pinn}}

Fix $\lambda$ and define the effective parameter
\[
    \vartheta
    =
    \theta_{\mathrm{eff}}(\lambda;\theta,\eta)
    =
    \theta + B(\lambda,\eta).
\]
The local quadratic loss is
\[
    \mathcal{L}_{\lambda}(\vartheta)
    =
    \frac12
    (\vartheta-\theta^\star)^\top
    H_\lambda
    (\vartheta-\theta^\star).
\]
Let
\[
    g
    =
    \nabla_\vartheta
    \mathcal{L}_{\lambda}(\vartheta).
\]
Since the loss is quadratic,
\[
    g
    =
    H_\lambda
    (\vartheta-\theta^\star).
\]

\paragraph{Regular PINN.}
For the regular PINN, the effective parameter is simply
\[
    \vartheta=\theta_{\mathrm{reg}} .
\]
The gradient-flow dynamics is
\[
    \dot{\theta}_{\mathrm{reg}}
    =
    -g .
\]
Therefore,
\[
    \frac{d}{dt}
    \mathcal{L}_{\lambda}(\theta_{\mathrm{reg}})
    =
    g^\top \dot{\theta}_{\mathrm{reg}}
    =
    -\|g\|^2 .
\]
Hence,
\[
    -
    \frac{d}{dt}
    \mathcal{L}_{\lambda}(\theta_{\mathrm{reg}})
    =
    \|g\|^2 .
\]

\paragraph{Loss-conditional PINN.}
For the loss-conditional PINN,
\[
    \vartheta
    =
    \theta + B(\lambda,\eta).
\]
For fixed $\lambda$, the gradients with respect to the trainable
variables $\theta$ and $\eta$ are obtained by the chain rule:
\[
    \nabla_\theta
    \mathcal{L}_{\lambda}
    =
    g,
\]
because $\partial \vartheta / \partial \theta$ is the identity, and
\[
    \nabla_\eta
    \mathcal{L}_{\lambda}
    =
    D_\eta B(\lambda,\eta)^\top g .
\]
The gradient-flow dynamics in the extended parameter space is therefore
\[
    \dot{\theta}
    =
    -g,
    \qquad
    \dot{\eta}
    =
    -
    D_\eta B(\lambda,\eta)^\top g .
\]
The loss decay along this flow is
\[
    \frac{d}{dt}
    \mathcal{L}_{\lambda}
    \bigl(
        \theta+B(\lambda,\eta)
    \bigr)
    =
    -
    \left\|
        \nabla_\theta
        \mathcal{L}_{\lambda}
    \right\|^2
    -
    \left\|
        \nabla_\eta
        \mathcal{L}_{\lambda}
    \right\|^2 .
\]
Substituting the expressions for the gradients gives
\[
    \frac{d}{dt}
    \mathcal{L}_{\lambda}
    \bigl(
        \theta+B(\lambda,\eta)
    \bigr)
    =
    -
    \|g\|^2
    -
    \left\|
        D_\eta B(\lambda,\eta)^\top g
    \right\|^2 .
\]
Therefore,
\[
    -
    \frac{d}{dt}
    \mathcal{L}_{\lambda}
    \bigl(
        \theta+B(\lambda,\eta)
    \bigr)
    =
    \|g\|^2
    +
    \left\|
        D_\eta B(\lambda,\eta)^\top g
    \right\|^2 .
\]
Since
\[
    -
    \frac{d}{dt}
    \mathcal{L}_{\lambda}(\theta_{\mathrm{reg}})
    =
    \|g\|^2,
\]
we obtain
\[
    -
    \frac{d}{dt}
    \mathcal{L}_{\lambda}
    \bigl(
        \theta+B(\lambda,\eta)
    \bigr)
    =
    -
    \frac{d}{dt}
    \mathcal{L}_{\lambda}(\theta_{\mathrm{reg}})
    +
    \left\|
        D_\eta B(\lambda,\eta)^\top g
    \right\|^2 .
\]
This proves the claimed identity.

Finally, since
\[
    g
    =
    H_\lambda(\vartheta-\theta^\star),
\]
we have
\[
    \left\|
        D_\eta B(\lambda,\eta)^\top g
    \right\|^2
    =
    \left\|
        D_\eta B(\lambda,\eta)^\top
        H_\lambda(\vartheta-\theta^\star)
    \right\|^2 .
\]
If
\[
    B(\lambda,\eta)
    =
    \sum_{j=1}^m \lambda_j M_j \eta,
\]
then
\[
    D_\eta B(\lambda,\eta)
    =
    \sum_{j=1}^m \lambda_j M_j .
\]
Hence the additional decay term becomes
\[
    \left\|
        \left(
            \sum_{j=1}^m \lambda_j M_j
        \right)^\top
        H_\lambda
        (\vartheta-\theta^\star)
    \right\|^2 .
\]
This completes the proof.

\subsection{Interpretation}

The result shows that, in the local quadratic regime, the conditional
architecture has a larger instantaneous descent capacity than the
regular PINN at the same effective parameter vector. The regular PINN
can move only through the original parameter direction $\theta$, whereas
the loss-conditional PINN can additionally move through the conditional
parameters $\eta$. These extra directions contribute the nonnegative
term
\[
    \left\|
        D_\eta B(\lambda,\eta)^\top
        \nabla_\vartheta
        \mathcal{L}_{\lambda}(\vartheta)
    \right\|^2
\]
to the instantaneous loss-decay rate.

This term vanishes only when the loss gradient is orthogonal to the
subspace generated by the conditional parameters. Otherwise, the
loss-conditional model decreases the same local quadratic loss strictly
faster at the same effective point.

The statement should be interpreted locally. It does not imply that
loss-conditioning changes the global optimum: in the well-specified
case, the common target remains $\theta^\star$. Rather, the theorem
shows that conditioning enriches the parameterization through which the
optimizer approaches this optimum. This provides a possible explanation
for the empirical observation that LC-PINN is less sensitive to a single
poor loss-weight choice and can explore a family of loss-weight
geometries during training.

\subsection{Remark on discrete gradient descent}

The theorem is stated for continuous-time gradient descent. For discrete
gradient descent with step size $\alpha$, the same comparison holds to
first order in $\alpha$. Indeed, for a smooth loss $F$, one step of
gradient descent gives
\[
    F(z-\alpha \nabla F(z))
    =
    F(z)
    -
    \alpha \|\nabla F(z)\|^2
    +
    O(\alpha^2).
\]
Applying this expansion to the regular and loss-conditional
parameterizations yields the same leading-order difference:
\[
    \alpha
    \left\|
        D_\eta B(\lambda,\eta)^\top
        \nabla_\vartheta
        \mathcal{L}_{\lambda}(\vartheta)
    \right\|^2
    +
    O(\alpha^2).
\]
Thus, for sufficiently small step sizes, the loss-conditional
parameterization has a larger one-step decrease whenever the additional
gradient component is nonzero.
\section{Hyperparameters and training details}
\label{sec:appendix-hyper}

This appendix collects the hyperparameters used for every method
in \Cref{sec:results}. All numbers are the configuration that
generated the headline tables in \Cref{sec:results}.

\paragraph{Shared LC-PINN configuration.}
Network: 4-layer MLP, hidden widths $[64, 64, 64, 64]$, $\tanh$
activations, Xavier initialisation. Conditioning route is
mode-dependent: concat at the input layer for the loss-weight mode
(Burgers, BL); FiLM \citep{perez2018film} on every hidden layer
followed by an L-BFGS finishing pass with a revert-on-worse safety
check for the parametric-coefficient mode (Helmholtz, Schrödinger).
Optimiser: Adam,
$\eta\!=\!10^{-3}$, cosine-annealed to $10^{-5}$ over the full
budget, gradient clip $1.0$. Mini-batch composition: $n_\lambda\!=\!4$
parameter samples per step, each with the per-equation collocation
counts below. Backend: PyTorch~2.9 with the MPS device on a single
Apple~M4 Max.

\paragraph{Per-equation collocation and budgets.}
\Cref{tab:hyper-lc} lists the LC-PINN collocation counts and step
budgets per PDE family.

\begin{table}[!t]
\centering
\caption{LC-PINN collocation counts and step budgets per PDE family.}
\label{tab:hyper-lc}
\begin{tabular}{lcccc}
\toprule
PDE & interior & boundary & initial & Adam / L-BFGS \\
\midrule
Burgers           & $1024$ & $200$ & $200$ & $50\text{k}\,/\,0$    \\
BL (viscous)      & $1024$ & $200$ & $200$ & $50\text{k}\,/\,0$    \\
Helmholtz (1D)    & $1024$ & $100$ & ---   & $50\text{k}\,/\,1.5\text{k}$ \\
Helmholtz (2D)    & $4096$ & $400$ & ---   & $25\text{k}\,/\,1.5\text{k}$ \\
Helmholtz (3D)    & $8192$ & $600$  & ---  & $25\text{k}\,/\,1.5\text{k}$ \\
Schrödinger (1D)  & $1024$ & $64$  & ---   & $50\text{k}\,/\,1.5\text{k}$ \\
\bottomrule
\end{tabular}
\end{table}

\paragraph{Baseline configurations.}
All single-$\lambda$ baselines share the LC-PINN backbone (same
widths, activation, initialisation, and Adam schedule) so that the
amortisation comparison is on equal architectural footing.

\begin{itemize}\setlength{\itemsep}{0pt}
\item \emph{SA-PINN \citep{mcclenny2020}}: per-collocation-point
      trainable weight ascended on the residual; Adam phase
      $10\text{k}$ iterations, then weights frozen and L-BFGS phase
      $5\text{k}$ iterations on the unweighted loss.
\item \emph{ReLoBRaLo \citep{bischof2021}}: 3-term loss-weight
      balancer (residual / boundary / initial) with softmax over
      log-loss-ratios, smoothed by EMA. Hyperparameters
      $\alpha\!=\!0.999$, $\tau\!=\!0.1$,
      $\rho_{\text{mean}}\!=\!0.999$.
\item \emph{Causal-PINN \citep{wang2024}}: time-causal residual mask
      $w(t)\!=\!\exp\!\left(-\varepsilon\,R(<\!t)\right)$ with
      $\varepsilon\!=\!100$. Burgers and BL only.
\item \emph{PI-DeepONet \citep{wang2021pi-deeponet}}: branch--trunk
      architecture; branch encodes the wavenumber $k$ via a 4-layer
      MLP (hidden width $128$); trunk encodes the spatial coordinate
      $x$ via a 4-layer MLP (hidden width $128$). Final solution is
      the inner product of branch and trunk outputs. Loss is the
      Helmholtz residual on the same collocation grid as the per-$k$
      LC-PINN, summed over an $n_k$-grid of branch inputs ($n_k\!=\!20$).
      Adam phase $50$k iterations followed by L-BFGS phase $1.5$k
      iterations with a revert-on-worse safety check; on 1D Helmholtz
      one seed (seed 1) diverged in L-BFGS and was reverted to the
      Adam-phase weights, reported as such in the 5-seed mean of
      \Cref{tab:helmholtz}.
\item \emph{FNO \citep{li2020fno}}: width $64$, $64$ Fourier modes,
      $4$ \texttt{SpectralConv1d}~+~\texttt{Conv1d}-shortcut blocks
      ($\sim\!1.07$M parameters). Optimiser: Adam,
      $\eta\!=\!10^{-3}$, weight decay~$10^{-4}$, \texttt{StepLR}
      with step~$=\!\text{epochs}/5$ and $\gamma\!=\!0.5$. Loss:
      per-sample rel-$L^2$. Training set: $512$ random Fourier
      initial conditions (truncated 6-mode spectrum, decaying
      amplitude) solved by the same Radau scheme used for the
      reference; $2000$ epochs.
\end{itemize}

\paragraph{Reference solutions.}
Burgers reference: Fourier-spectral spatial discretisation with a
Radau IIA implicit time integrator, integrated to relative residual
$<\!10^{-6}$ on a $512\!\times\!1001$ space--time grid.
Buckley--Leverett reference: explicit Lax--Friedrichs flux on
$n_x\!=\!1000$ with centred-difference viscous regularisation,
CFL-limited time stepping. Helmholtz~1D reference: closed-form
manufactured solution $u(x;k)\!=\!\sin(\pi x)\cos(kx)$ with the
forcing $f$ chosen to match. Helmholtz~2D reference: closed-form
manufactured solution
$u(x,y;k) \!=\! \sin(\pi x)\sin(\pi y)\cos(kx)\cos(ky)$. Letting
$a(x;k)\!=\!\sin(\pi x)\cos(kx)$ and
$b(y;k)\!=\!\sin(\pi y)\cos(ky)$, so $u\!=\!a(x;k)\,b(y;k)$, the
forcing
\[
   f(x,y;k) \;=\; \Delta u + k^2 u
   \;=\; -\bigl(2\pi^2 + k^2\bigr)\, u
         \;-\; 2\pi k\,\bigl[\cos(\pi x)\sin(kx)\sin(\pi y)\cos(ky) +
                              \sin(\pi x)\cos(kx)\cos(\pi y)\sin(ky)\bigr]
\]
agrees with a centred-difference numerical Laplacian on the
manufactured solution to absolute error $<\!4\!\times\!10^{-6}$.
Helmholtz~3D reference: closed-form manufactured solution
$u(x,y,z;k)\!=\!\sin(\pi x)\sin(\pi y)\sin(\pi z)\cos(kx)\cos(ky)\cos(kz)$
on the unit cube, with the matching forcing
$f \!=\! \Delta u + k^2 u$ derived analytically; agreement to a
centred-difference numerical Laplacian is $<\!10^{-5}$ across
$k\!\in\!\{1,3,5\}$. Schrödinger reference: with $u(x;\alpha)\!=\!\sin(\pi x)\,g(x;\alpha)$,
$g(x;\alpha)\!=\!e^{-\alpha (x-1/2)^2/2}$, the forcing is
\[
   f(x;\alpha) \;=\; -u'' + V(x;\alpha)\, u
   \;=\; (\pi^2 + \alpha)\, u
         \;+\; 2\pi\alpha\,(x-\tfrac12)\,\cos(\pi x)\, g(x;\alpha),
\]
which agrees with a centred-difference numerical residual to
$<\!5\!\times\!10^{-6}$ across $\alpha\in\{0.5,2,5,10\}$.

\paragraph{Evaluation grids.}
Burgers / BL test errors are computed on a $256\!\times\!101$
space--time grid; Helmholtz on a $1024$-point spatial grid. The
$K$-shot averaged LC-PINN prediction is the mean of $K$ forward
passes at independent $\lambda$ samples from $p_\lambda$; we use
$K\!=\!100$ for Burgers and $K\!=\!25$ for BL in the headline
tables.

\paragraph{Compute.}
Each run uses a single Apple~M4~Max with the PyTorch MPS backend.
Reported wall times are end-to-end per seed (model construction,
training, and evaluation), measured with \texttt{time.perf\_counter}.

\paragraph{Reproducibility.}
Random seeds: $\{0, 1, 2, 3\}$ for the four headline replicates;
$\{0, 1\}$ for the 2-seed ablation budgets. Each seed controls the
PyTorch global RNG and the NumPy RNG used to draw collocation
points and $\lambda$-samples. The full training script, equation
modules, and reference solvers are released with the
camera-ready version.

\section{Ablations: full protocol and numbers}
\label{sec:appendix-ablations}

\paragraph{$K$-shot averaging.}
Sweeping $K \in \{25, 50, 100, 200, 400\}$ on the trained LC-PINN
gives essentially constant rel-$L^2$ ($3.51\!\times\!10^{-3}$ at
$K\!=\!25$, $3.49\!\times\!10^{-3}$ at $K\!=\!100$,
$3.49\!\times\!10^{-3}$ at $K\!=\!400$; spread
$\pm 2.7\!\times\!10^{-3}$ stays roughly constant). At the optimum
each $\lambda$-slice is already a residual minimiser
(Proposition~1), so increasing $K$ adds samples but no new
information.

\paragraph{$n_\lambda$ per training step.}
At a matched 25k-epoch budget on Burgers (2 seeds), sweeping
$n_\lambda \!\in\! \{1, 4, 16\}$ per step gives non-monotone
behaviour: $n_\lambda\!=\!1\!\to\!1.69\!\times\!10^{-2}$,
$n_\lambda\!=\!4\!\to\!1.22\!\times\!10^{-2}$,
$n_\lambda\!=\!16\!\to\!1.50\!\times\!10^{-1}$. Too few
$\lambda$ per step makes the gradient estimate of the
$p_\lambda$-expectation noisy (Remark~3 / Hoeffding); too many
concentrates compute into fewer optimisation steps and the network
does not converge in the fixed budget. At the headline 50k-epoch
budget we use $n_\lambda\!=\!4$.

\paragraph{Sampling distribution $p_\lambda$.}
At matched 25k-epoch / $n_\lambda\!=\!4$ / 2-seed budget on
Burgers, swapping $p_\lambda$ from uniform on the simplex to
log-uniform gives rel-$L^2$ of
$1.22\!\times\!10^{-2} \pm 5.14\!\times\!10^{-3}$ (uniform) versus
$1.74\!\times\!10^{-1} \pm 9.26\!\times\!10^{-3}$ (log-uniform), a
$\sim\!14\times$ degradation. We read this not as a contradiction
with Remark~2 (an asymptotic-optimum statement) but as a
finite-budget effect: log-uniform places dominant mass on small
loss weights, weakening the gradient signal at large $\lambda$
values where the residual contributes most to the test-relevant
slices.

\section{FiLM does not help in loss-weight mode}
\label{sec:appendix-film-loss-weight}

In the parametric-coefficient mode (Helmholtz, Schrödinger) FiLM
modulation lifts the LC-PINN by an order of magnitude: on 1D
Helmholtz, plain concat reaches rel-$L^2 \!\sim\!9\!\times\!10^{-3}$
on the same $k$-grid (\texttt{lc\_pinn\_helmholtz} runs), whereas
FiLM$+$L-BFGS reaches $9.4\!\times\!10^{-4}$ (\Cref{tab:helmholtz},
$\sim\!10\times$ tighter). In the loss-weight mode (Burgers) it
regresses: a four-seed re-run of Burgers with
FiLM$+$L-BFGS gave $5.7\!\times\!10^{-2} \pm 5.5\!\times\!10^{-2}$,
with two seeds at the concat baseline ($\sim\!3\!\times\!10^{-3}$)
and two trapped near $10^{-1}$. We read this as evidence that
FiLM's value lies in disentangling \emph{operator}-changing inputs
from spatial coordinates; loss-weight inputs do not change the
operator, so FiLM gates over a degree of freedom that does not
exist. The headline Burgers and BL rows in
\Cref{sec:results-burgers-bl} therefore use the simpler
concat$+$Adam configuration of \citet{dosovitskiy2020}.

\section{Per-parameter breakdowns}
\label{sec:appendix-per-param}

\begin{figure}[H]
\centering
\includegraphics[width=0.85\linewidth]{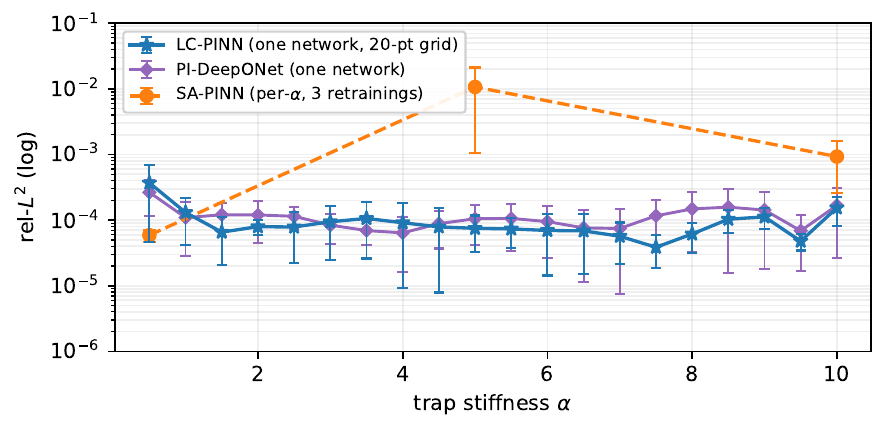}
\caption{Per-$\alpha$ rel-$L^2$ on 1D parametric Schrödinger.
LC-PINN (blue, 4 seeds) and PI-DeepONet (purple, 4 seeds) cover the
20-point grid from one trained network each; SA-PINN (orange, 2
seeds) is three separate retrainings at $\alpha\in\{0.5,5.0,10.0\}$.
Error bars are $\pm$std on a log axis.}
\label{fig:per_alpha_schrodinger}
\end{figure}

The per-$\alpha$ breakdown for 1D Schrödinger
(\Cref{sec:results-schrodinger}) is given in
\Cref{tab:schrod-per-alpha} and visualised in
\Cref{fig:per_alpha_schrodinger}. The two
amortised methods (LC and DON) hold a flat $\sim\!10^{-4}$ curve
across the entire $\alpha\in[0.5,10]$ range; the per-$\alpha$
SA-PINN baseline matches at $\alpha=0.5$ and $\alpha=10$ but
collapses by two orders of magnitude at the intermediate
$\alpha=5.0$ training point --- the same incoherence pattern
documented for Helmholtz in \Cref{sec:results-helmholtz}.

\begin{table}[!t]
\centering
\caption{Per-$\alpha$ rel-$L^2$ on 1D parametric Schrödinger
(mean $\pm$ std over 4 LC seeds and 2 SA seeds per $\alpha$).
Bold: best per row.}
\label{tab:schrod-per-alpha}
\small
\begin{tabular}{r|cc}
\toprule
$\alpha$ & SA-PINN (per-$\alpha$) & LC-PINN (one net) \\
\midrule
$0.5$  & $\mathbf{(5.89\!\pm\!1.26)\!\times\!10^{-5}}$ & $(3.69\!\pm\!3.23)\!\times\!10^{-4}$ \\
$5.0$  & $(1.07\!\pm\!1.05)\!\times\!10^{-2}$ & $\mathbf{(7.50\!\pm\!4.25)\!\times\!10^{-5}}$ \\
$10.0$ & $(9.29\!\pm\!6.67)\!\times\!10^{-4}$ & $\mathbf{(1.54\!\pm\!0.72)\!\times\!10^{-4}}$ \\
\midrule
\textbf{grid mean} & $3.89\!\times\!10^{-3}$ & $\mathbf{1.99\!\times\!10^{-4}}$ \\
\bottomrule
\end{tabular}
\end{table}

\end{document}